\documentclass[runningheads]{llncs}

\usepackage[T1]{fontenc}
\usepackage{graphicx}
\usepackage{amsmath,amssymb}
\usepackage{booktabs}
\usepackage{multirow}
\usepackage{tabularx}
\usepackage[protrusion=true,expansion=false]{microtype}
\usepackage[table]{xcolor}
\usepackage{pgfplotstable}
\definecolor{bestRed}{HTML}{F8CECC}
\definecolor{worstYellow}{HTML}{FFFFBF}
\newcommand{\cc}[1]{\cellcolor{bestRed!#1!worstYellow}}
\usepackage{tikz}
\usepackage{float}
\usetikzlibrary{arrows.meta, positioning, shapes.geometric, fit, calc}
\usepackage{url}
\usepackage{soul}

\begin{document}

\title{FOSCo-Net: Ego-centric Voxel Completion for Forest LiDAR Occlusions}

\titlerunning{FOSCo-Net: Ego-centric Voxel Completion for Forest LiDAR Occlusions}

\author{Iman Esfandiyar\inst{1,2}\orcidID{0000-0002-3360-7728} \and
        Dominik Belter\inst{1}\orcidID{0000-0003-3002-9747}}

\authorrunning{I. Esfandiyar and D. Belter}

\institute{Poznań University of Technology,
  Institute of Robotics and Machine Intelligence, Poznań, Poland \and
  Łukasiewicz Research Network -- Poznań Institute of Technology,
  Center of Digital Transformation,
  Department of Artificial Intelligence, Poznań, Poland\\
  \email{iman.esfandiyar@put.poznan.pl}
}

\maketitle
\vspace{-0.5cm}
\begin{abstract}
Ground robots operating in temperate forests suffer from systematic LiDAR occlusions caused by tree trunks, creating persistent blind regions that degrade traversability estimation. In this paper, we propose FOSCo-Net, a self-supervised voxel completion framework that learns to reconstruct occluded geometry from a single scan using temporally accumulated observations as automatic supervision, and we evaluate cylindrical and Cartesian ego-centric grids with multiple decoder designs. FOSCo-Net consistently improves reconstruction quality and traversal safety, with a gated skip decoder achieving the most reliable performance and significantly reducing false obstacle predictions compared to baselines.

\end{abstract}

\section{Introduction}
\label{sec:intro}

Autonomous robots operating in forests must maintain accurate
traversability estimates to navigate safely around trees and undergrowth.
Spinning LiDAR sensors are the primary exteroceptive sensor for this
task, but they suffer from persistent \emph{occlusion shadows}: every
trunk blocks the sensor rays behind it, leaving wedge-shaped unobserved
volumes that are recorded as unknown space in occupancy maps.
Unlike random sensor noise, these shadows are \emph{systematic}: the same trunk consistently blocks the same angular sector relative to the sensor heading.

Occlusion shadows have safety-critical consequences for motion planning. Planners typically treat unknown space as either free (risking collision with hidden obstacles) or occupied (blocking traversable corridors behind trunks)~\cite{roldao2020lmscnet}. In dense forests, the conservative policy fragments the free-space map into disconnected islands, often preventing the planner from finding any path. By predicting occluded geometry \emph{before} planning, the proposed FOSCo-Net converts unknown voxels into confident occupancy or free estimates, giving the planner a complete local map for safe routing through the corridor behind each trunk.

\subsection{Related Work}
\label{sec:related}
Semantic scene completion (SSC) aims to infer full 3D geometry from partial observations, originally targeting camera-based inputs with large annotated datasets. LMSCNet~\cite{roldao2020lmscnet} extends SSC to single-frame LiDAR using a coarse-to-fine voxel decoding strategy. More recent generative approaches such as DiffSSC~\cite{Cao2025} use denoising diffusion probabilistic models (DDPMs) to iteratively corrupt and reconstruct data, enabling high-quality scene completion from noise. In parallel, S3CNet~\cite{Cheng2021} combines sparse 2D bird’s-eye-view (BEV) and 3D voxel representations with dedicated completion networks and BEV-guided voxel fusion.

Other methods~\cite{Tao2022,Boettcher2023} improve sparse LiDAR depth using overlapping camera views to obtain denser reconstructions and better free-space estimates, but they rely on visible or multi-view cues and do not explicitly reconstruct regions occluded by foreground obstacles. Self-supervised alternatives reduce annotation requirements by exploiting temporal consistency: Vizzo et al.~\cite{vizzo2022makedense} use accumulated scans as pseudo-ground truth for TSDF-based completion, but their formulation is world-centric and less suited for pose-invariant online inference. More recently, implicit neural map representations reconstruct large-scale scenes from full posed scan sequences: SHINE-Mapping~\mbox{\cite{zhong2023shine}} fits a sparse hierarchical feature grid to a scan sequence offline, and 4dNDF~\mbox{\cite{zhong2024ndf}} extends this formulation with a time-dependent TSDF that additionally separates dynamic objects. These approaches deliver accurate reconstructions of \emph{observed} space, but they require the complete sequence and per-scene test-time optimization, and they cannot infer geometry in regions never covered by any sensor ray. Our task is complementary: predicting \emph{unobserved}, occluded space from a single scan in one forward pass of a network trained once, which rules out such sequence-based mapping methods as direct baselines. In contrast, world-frame representations support large-scale mapping~\cite{xujia2025} but are computationally heavier, while ego-centric representations are more efficient for local control~\cite{liang2024}, including BEV fusion~\cite{mei2023} and top-view prediction~\cite{gu2024}. Finally, sparse convolutional networks~\cite{yan2020sparsity} enable efficient processing by operating only on occupied voxels, while learned skip-gating mechanisms~\cite{zioulis2022hybrid} reduce false positives by suppressing unreliable features. These ideas motivate our design, including a sparse encoder and $\sigma$-gated skip connections that explicitly handle unobserved regions.

To the best of our knowledge, no prior work addresses the systematic trunk
occlusions of forest environments with a completion network that is simultaneously
(i)~self-supervised, with supervision derived from robot motion alone and no manual
labels, (ii)~ego-centric, hence pose-invariant and applicable per scan without a
persistent world map, and (iii)~evaluated on traversability safety rather than
reconstruction quality alone: camera-based SSC relies on annotated datasets,
LMSCNet-style LiDAR completion is trained against world-frame ground truth, and the
self-supervision of Vizzo et al.~is world-centric and map-bound. FOSCo-Net closes
this gap.

Our contribution can be summarized as follows:
\begin{itemize}
\item to our knowledge, the first self-supervised, ego-centric
occlusion-completion framework for forest LiDAR, in which the supervision target
\mbox{(Eq.~\ref{eq:target})} is derived automatically from robot motion, with no
annotations and no world map at inference,

\item a new traversability-oriented evaluation metric, SFPR, which measures
false obstacles in the occluded corridor behind each trunk, the region that
determines whether a planner can route through the corridor at all,

\item an asymmetric $\sigma$-gated skip decoder whose gates vanish exactly at
unobserved positions, anchoring predictions to observed surfaces, validated by a
controlled ablation of four decoders on two ego-centric grids (cylindrical and
Cartesian) which shows that decoder design, not grid choice, drives performance.
\end{itemize}
\section{Problem Formulation}
\label{sec:dataset}

\subsection{Scene Occlusion in Forest Environment}
\begin{figure}[htb]
  \centering
  \includegraphics[width=0.6\linewidth]{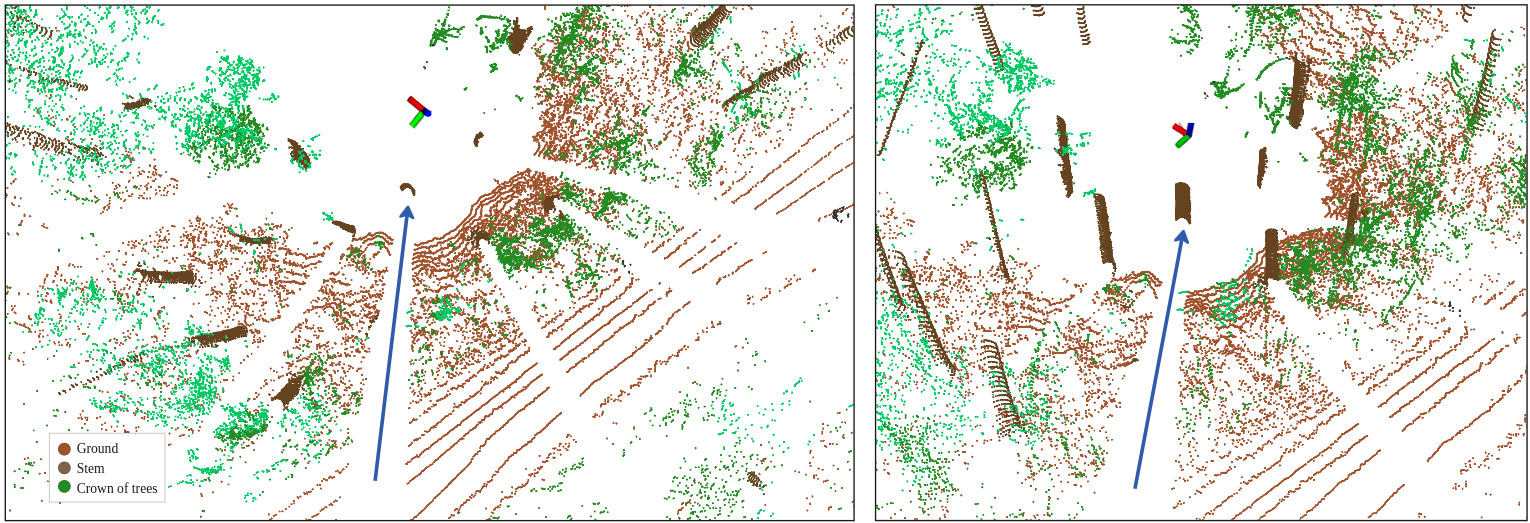}
  \caption{Single LiDAR scan in a temperate forest.
           The bright gap (blue arrow) behind a tree trunk is an
           occlusion shadow.}
  \label{fig:occlusion}
\end{figure}

As a robot traverses a forest, each LiDAR sweep captures only the geometry \emph{visible} from the current sensor pose (Fig.~\ref{fig:occlusion}).
Tree trunks cast persistent occlusion shadows: angular sectors behind
every trunk contain no measurements regardless of how many sweeps are accumulated at the same location. Considering the entire sequence of LiDAR motions, these shadows may already be visible from previous viewpoints. This property enables scan completion methods to learn the missing geometry from neighboring observations.

\subsection{Self-Supervised Methodology}
\begin{figure}[htb]
  \centering
  \includegraphics[width=0.6\linewidth]{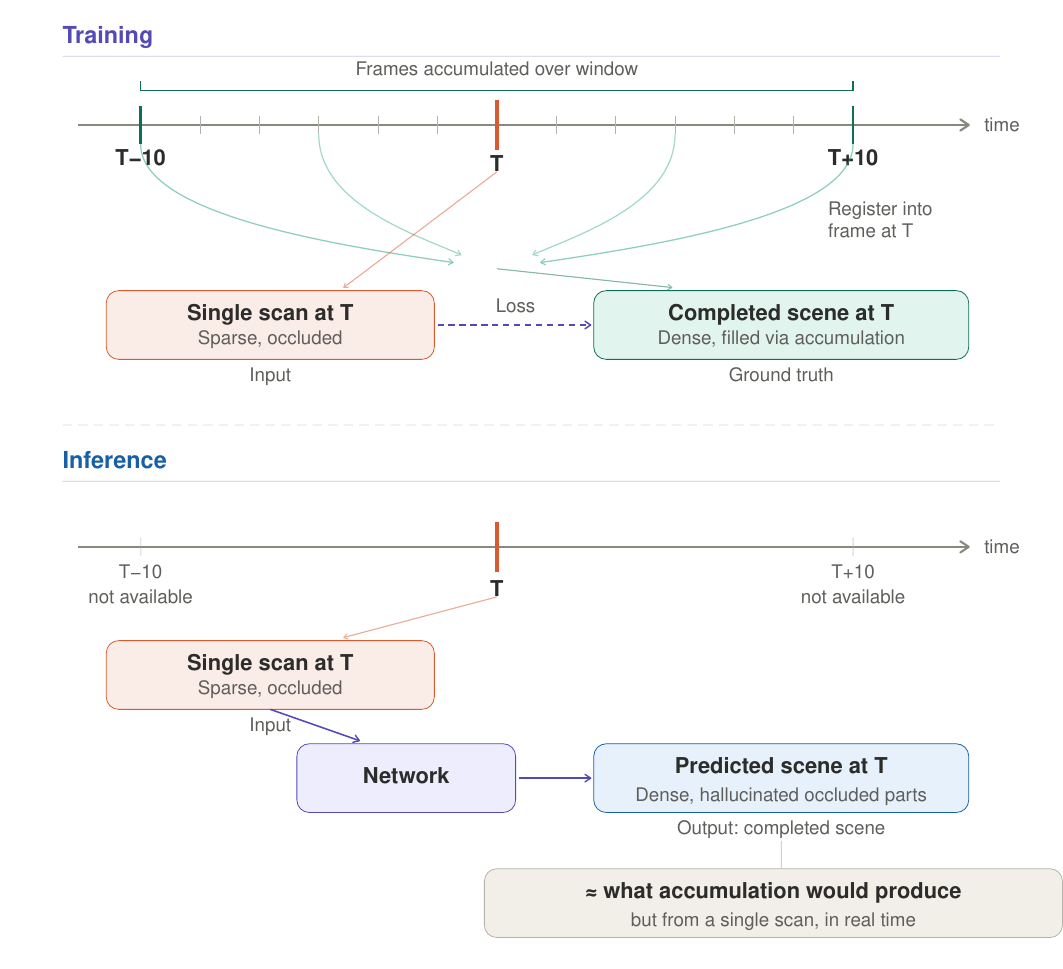}
  \caption{Self-supervised training and inference pipeline.
           \textbf{Top:} LiDAR frames from a temporal window
           $[T{-}\Delta,\,T{+}\Delta]$ are registered into the
           frame at~$T$ to form the dense supervision target.
           \textbf{Bottchaom:} at inference only the single scan
           at~$T$ is available.}
  \label{fig:pipeline}
\end{figure}

By registering and merging scans from a surrounding temporal window \\
$[T{-}\Delta,\,T{+}\Delta]$ into the coordinate frame at time~$T$,
we obtain a substantially more complete picture of the local scene
(Fig.~\ref{fig:pipeline}).

Let $\mathcal{S}_t$ denote the occupied voxels in the single scan
and $\mathcal{M}_t$ those in the accumulated map.
The training target is the set difference
\begin{equation}
  \mathcal{T}_t = \mathcal{M}_t \setminus \mathcal{S}_t ,
  \label{eq:target}
\end{equation}
i.e.\ voxels present in the map but absent from the current scan,
exactly the occluded voxels the network must predict.
We use $N{=}20$ accumulated frames.
This \emph{requires no human annotation}: the supervision signal is
derived automatically from robot motion.

\section{Voxel Grid Representations}
\label{sec:grids}

Both grid representations proposed in this paper are ego-centric: raw world-frame points are
transformed to a sensor-centred, yaw-aligned frame so the network
always sees the scene from the same canonical viewpoint regardless of
the robot's absolute pose.
We compare two discretisations of this frame, keeping the cell size
$\Delta{=}0.1$\,m and the vertical extent $[-5, 20]$\,m identical in
both cases, yielding the same total cell count of 3.6M.

\subsection{Cylindrical Grid}
\label{sec:cylindrical}
Points in the ego frame are converted to cylindrical coordinates
$(r, \theta, z)$ and discretised.
Grid parameters are in Table~\ref{tab:grid}.
The radial range 2–6\,m covers the trunk cross-section region while
excluding the sensor blind-spot and distant points.
\begin{table}[b]
  \caption{Grid parameters for both representations ($^*$cylindrical only).
           Both grids share the same vertical extent and 0.1\,m
           resolution, yielding 3.6M total cells.}
  \label{tab:grid}
  \centering
  \begin{tabular}{lcccc}
    \toprule
    \textbf{Dimension} & \textbf{Cylindrical} & \textbf{  Cartesian}
                       & \textbf{   Range} & \textbf{Resolution} \\
    \midrule
    Radial $r$ / $x$   & 40  & 120 & 2–6\,m$^*$ / $\pm$6\,m & 0.10\,m \\
    Angular $\theta$ / $y$ & 360 & 120 & 0°–360°$^*$ / $\pm$6\,m & 1°$^*$ / 0.10\,m \\
    Vertical $z$        & 250 & 250 & $-$5–20\,m & 0.10\,m \\
    \midrule
    \multicolumn{3}{l}{\textbf{Total cells}} & \multicolumn{2}{r}{3,600,000} \\
    \bottomrule
  \end{tabular}
\end{table}
Each occupied voxel carries a three-channel feature vector

$\mathbf{f} = [1,\;\mathbf{1}(\ell{=}\textsc{Ground}),\;
\mathbf{1}(\ell{=}\textsc{Stem})]^\top$,
encoding occupancy and semantic class. The $\theta$ dimension is angularly periodic: $\theta{=}0$ and
$\theta{=}360^\circ$ are physical neighbours. We apply \textbf{circular padding} on $\theta$ before every dense
Conv3d so that $3{\times}3{\times}3$ kernels see a contiguous
neighbourhood across the wrap-around boundary. 

\subsection{Cartesian Grid}
\label{sec:cartesian}

The Cartesian grid discretises the ego frame directly in $(x, y, z)$
with uniform 0.1\,m cells covering $x,y \in [-6, +6]$\,m and
$z \in [-5, +20]$\,m (Table~\ref{tab:grid}).
The $x$-axis points in the sensor's forward direction (yaw-aligned),
preserving the same pose-invariant property as the cylindrical grid.
Feature encoding is identical ($\mathbf{f}$ as above).
Unlike the cylindrical grid, all axes are spatially uniform: standard
zero-padded convolutions apply without any circular wrapping, and
$3{\times}3{\times}3$ kernels see symmetric Euclidean neighborhoods.

\begin{figure}[t]
  \centering
  \includegraphics[width=1.0\linewidth]{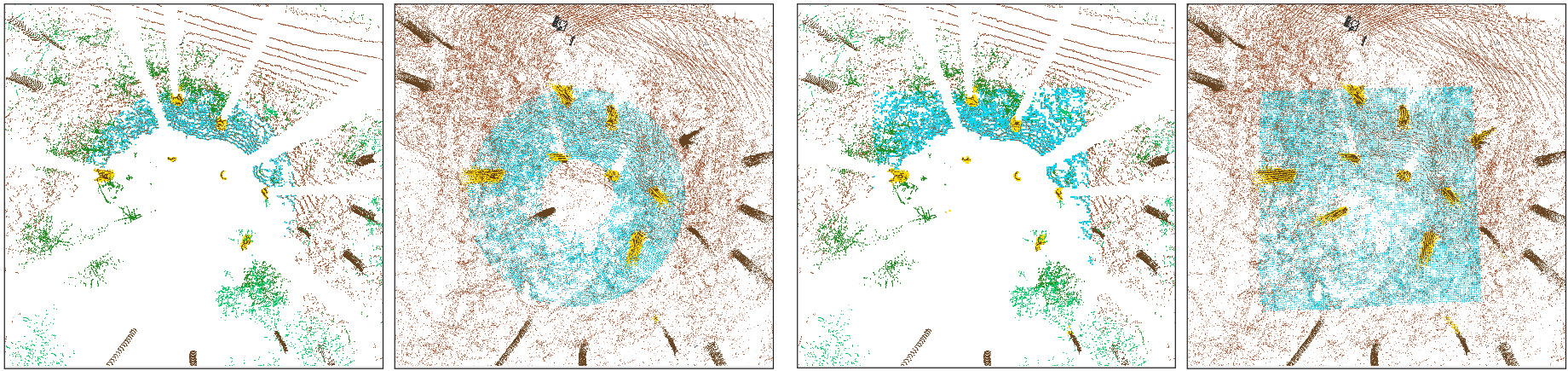}
  \put(-345,75){a} \put(-259,75){b} \put(-169,75){c} \put(-83,75){d}
  \caption{Comparison between Ego-centric grid pairs: cylindrical input (cyan\,=\,ground, yellow\,=\,stem, brown and green points are not used by both representation) (a) and target output (b),Cartesian input (c) and target output (d). The model predicts voxels in the target absent from the input.}
  \label{fig:grids}
\end{figure}

The differences between the cylindrical and cartesian grid are presented in Fig.~\ref{fig:grids}. 

\section{Model Architectures}
\label{sec:architectures}

Each design decision follows from a property of the problem. Occlusion shadows
are systematic, so their geometry is predictable from the visible surfaces that cast
them, LiDAR returns exist only on surfaces, so the encoder computes features only
where observational evidence exists (sparse convolutions), and the characteristic
failure of dense decoders is hallucination far from any observation, so the skip
gates are constructed to vanish exactly at unobserved positions. We propose four FOSCo-Net decoder variants which share a common sparse encoder and produce a dense logit tensor for every cell in the active
grid,\\ $(B, 1, 40, 360, 250)$ for cylindrical and
$(B, 1, 120, 120, 250)$ for Cartesian.

\subsection{Shared Sparse Encoder}

The encoder uses submanifold sparse convolutions to process only the
${\approx}0.7\%$ occupied input voxels.
Three SparseResBlocks (Enc0--2, $C$/$2C$/$4C$ channels) alternate with
stride-2 \texttt{SparseConv3d} downsampling, followed by a bottleneck at
$8C$ channels. With $C{=}16$, the bottleneck is $5{\times}45{\times}32$
(cylindrical) or $15{\times}15{\times}32$ (Cartesian), a $512{\times}$
reduction.

\begin{figure}[H]
  \centering
  \includegraphics[width=0.65\linewidth]{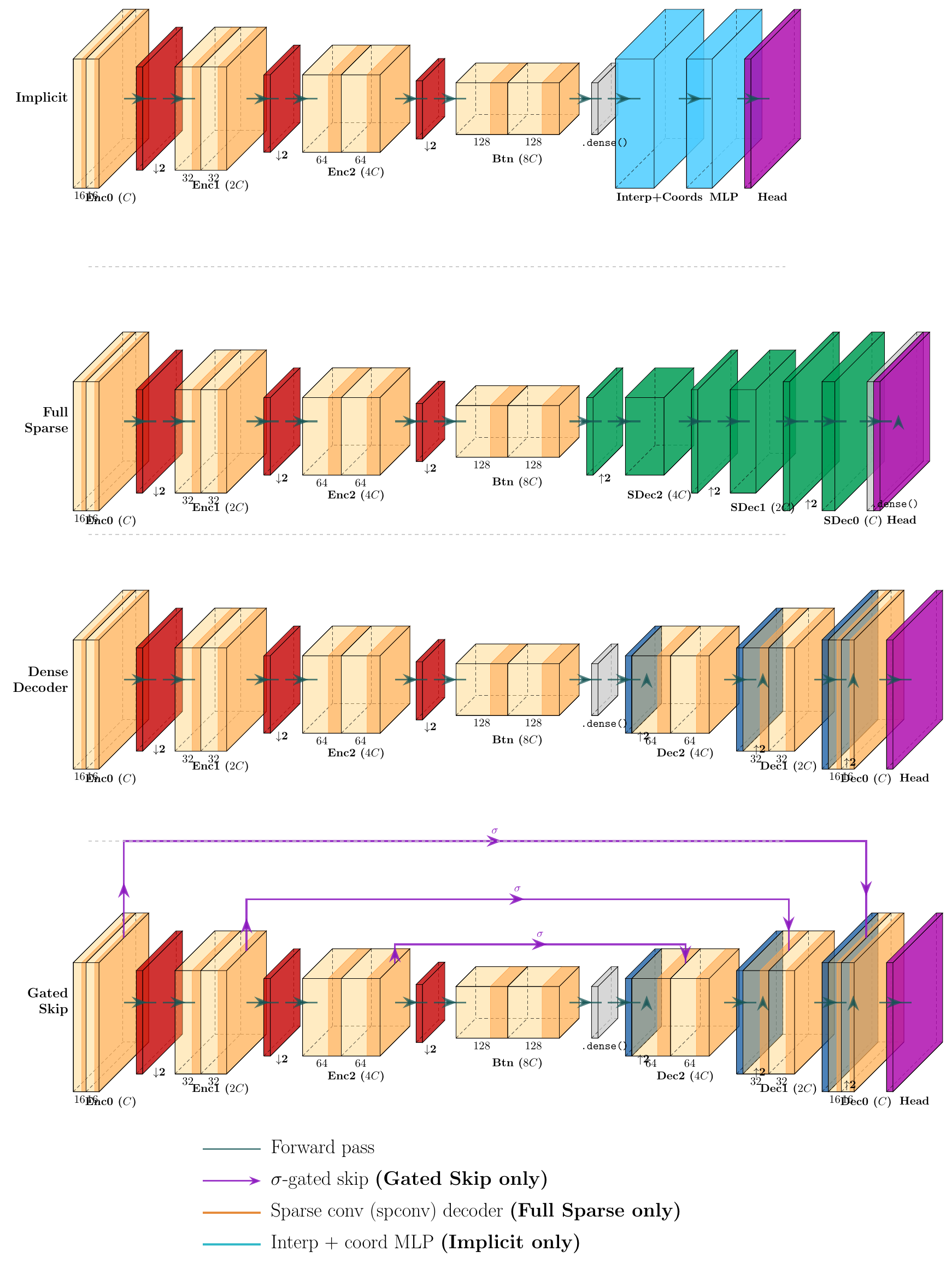}
  \put(-271,270){a} \put(-271,190){b} \put(-271,130){c} \put(-271,60){d}
  \caption{Proposed decoders based on the same sparse encoder (yellow/banded blocks): implicit, with a coordinate-conditioned $1{\times}1{\times}1$
MLP head (teal) (a), full Sparse, using \texttt{SparseConvTranspose3d}
upsamplers (orange) (b), dense Decoder, with trilinear DenseUpBlocks after an early \texttt{.dense()} (c), Gated Skip, extending the Dense Decoder with $\sigma$-gated skip connections (purple arcs) from each encoder level (d).}
  \label{fig:arch_all4}
\end{figure}

\subsection{Dense Decoder: Early Densification}

A single \texttt{.dense()} call converts the bottleneck to a dense
tensor (Fig.~\ref{fig:arch_all4}c).
Three DenseUpBlocks upsample via trilinear interpolation followed by
two $3{\times}3{\times}3$ Conv3d layers (circularly padded on $\theta$
for cylindrical, standard zero-padding for Cartesian) to the full output
resolution.
A $1{\times}1{\times}1$ head produces the occupancy logit.

\subsection{Full Sparse Decoder}

The Full Sparse decoder (Fig.~\ref{fig:arch_all4}b) stays
sparse throughout, with
\\ \texttt{SparseConvTranspose3d}($k{=}2$,$s{=}2$) expanding each active
voxel into $2^3{=}8$ outputs and growing the active set into
previously empty cells. A final \texttt{.dense()} and spatial crop
after three upsampling steps yield the output logits. Predictions
remain within kernel reach of observed voxels, giving a conservative
spatial prior. 

\subsection{Implicit Coordinate-Conditioned Decoder}

The Implicit decoder (Fig.~\ref{fig:arch_all4}a) densifies
the bottleneck and trilinearly upsamples it to full output resolution.
A normalised coordinate grid $\mathbf{c} \in [0,1]^3$ is appended
channel-wise, a stack of $1{\times}1{\times}1$ convolutions decodes
each position independently:
\begin{equation}
  \hat{y}(\mathbf{x})
    = \mathrm{MLP}\!\bigl(
        f_{\mathrm{btn}}^{\mathbf{x}},\;
        \tfrac{x_1}{N_1},\;\tfrac{x_2}{N_2},\;\tfrac{x_3}{N_3}
      \bigr),
    \quad\text{MLP: }8C{+}3 \to 64 \to 32 \to 1 .
\end{equation}
Coordinate channels supply a spatial prior, but the lack of decoder
convolutions removes circular-padding and local-context effects,
making this a useful comparison point. 

\subsection{Gated Skip Connections}

The Gated Skip architecture (Fig.~\ref{fig:arch_all4}d) adds
learned skip connections from each encoder level to the matching decoder
level in a U-Net style:
\begin{equation}
  d_i \leftarrow d_i
    + \sigma\!\bigl(\mathrm{Conv}_{1\times1}(d_i)\bigr)
      \cdot e_i^{\mathrm{dense}} .
  \label{eq:gate}
\end{equation}
At unobserved positions, $e_i^{\mathrm{dense}}{=}\mathbf{0}$ causes the gating term to vanish, allowing the decoder to operate without encoder guidance. This asymmetric formulation directly resolves the inherent ambiguity of the Dense Decoder: the gating mechanism explicitly indicates the location of the observed surface, thereby facilitating inference of the occluded geometry immediately adjacent to it. Notably, the proposed gating mechanism generalises without modification across both grid representations. 

\section{Training Protocol}
\label{sec:training}

All eight models (four architectures $\times$ two grids) are trained
identically in terms of loss, optimiser, and schedule, so any
difference in evaluation metrics reflects only the combination of
grid representation and decoder design.

The task is binary classification on each of the 3.6M cells.
With ${\approx}0.7\%$ positive rate, focal binary
cross-entropy~\cite{lin2017focal} with additional per-voxel weighting
compensates for class imbalance:
\begin{equation}
  \mathcal{L}
  = \frac{1}{|V|} \sum_{v \in V}
    w_v \,(1 - p_t^v)^\gamma\,
    \mathrm{BCE}(\hat{y}_v,\, Y_v) ,
\end{equation}
where $\gamma{=}2$, the base positive weight $w^+{=}20$, and the
additional stem weight $w_{\text{stem}}{=}8$.
Here
$\mathrm{BCE}(\hat{y}_v, Y_v) =
-\bigl[\,Y_v \log \hat{y}_v + (1{-}Y_v)\log(1{-}\hat{y}_v)\,\bigr]$
is the binary cross-entropy between the predicted occupancy probability
$\hat{y}_v \in (0,1)$ and the binary target $Y_v \in \{0,1\}$ from
\mbox{Eq.~(\ref{eq:target})}, and
$p_t^v = Y_v \hat{y}_v + (1{-}Y_v)(1{-}\hat{y}_v)$ is the probability
the model assigns to the true class of voxel~$v$.
Trunk voxels thus receive $160{\times}$ the loss weight of a negative
voxel, concentrating learning on the safety-critical class. We extract \textbf{50 ego-centric sample pairs} from a single sequence
(40 training, 10 Testing) of the DigiForests dataset~\cite{malladi2025digiforests} to train our network. 
Only \textsc{Ground} and \textsc{Stem} voxels are retained.
Identical samples are used for both grid variants, guaranteeing a
fair comparison. Small number of training samples requires strong data augmentation to provide generalization capabilities of the obtained neural network. Augmentations are applied jointly to scan and target so that the set
difference $\mathcal{T}_t$ is preserved. 

\textbf{Cylindrical:}
(i)~$\theta$-shift: uniform random rotation of all $\theta$ indices
by $\delta{\in}[0,359]$, exploiting circular periodicity;
(ii)~$\theta$-flip: mirroring $\theta_i{\leftarrow}N_\theta{-}1{-}\theta_i$
with probability 0.5;
(iii)~feature dropout ($p{=}0.05$).
\textbf{Cartesian:}
$\theta$-shift and flip are not applicable (Cartesian rotation would
misalign the grid); instead,
(i)~$x$-flip: mirroring $x_i{\leftarrow}N_x{-}1{-}x_i$ with
probability 0.5;
(ii)~$y$-flip: mirroring $y_i{\leftarrow}N_y{-}1{-}y_i$ with
probability 0.5;
(iii)~feature dropout ($p{=}0.05$).

All models are trained for \textbf{50 epochs} with AdamW
($\eta{=}10^{-3}$, weight decay $10^{-4}$) and a one-cycle
learning-rate schedule.
Batch size is 2 for Dense Decoder, Gated Skip, and Full Sparse on
both grids, 1 for Implicit on both grids .
Training takes approximately 25–30 minutes per model on an
NVIDIA RTX~4090\ (24\,GB).

\section{Experiments}
\label{sec:experiments}

We evaluate all ten models on held-out testing data, with emphasis on stem recall and traversability safety. All ten models are evaluated using six metrics: Loss, IoU, F$_1$, stem recall (Rec), global precision (Prec), and shadow false positive rate SFPR, a metric we introduce to quantify
traversability safety in the occluded corridor behind each trunk. Stem recall is the primary completeness target, SFPR is the primary
traversability target. SFPR measures false occupancy predictions in the radial corridor behind
each observed stem: letting $C_s$ denote the free-space voxels from the
stem's range to $r_{\max}$ in the same angular sector,
$\mathrm{SFPR} = |\{v{\in}C_s : \hat{y}_v{=}1, Y_v{=}0\}|\,/\,|C_s|$;
a high value means the planner is blocked from routing behind trunks.

\subsection{Results and Comparison}

We compare our approach against two dense-input baselines (no sparse encoder), each
on its native grid.
Both baselines share our problem setting, single-scan input and dense per-voxel
occupancy output, enabling a like-for-like comparison under identical training data,
loss, and metrics, sequence-based implicit mapping
methods~\mbox{\cite{zhong2023shine,zhong2024ndf}} require the full posed sequence at
inference and are therefore not applicable as single-scan predictors
\mbox{(cf.~Sec.~\ref{sec:related})}.
\textbf{MIDUNet2D}~\cite{vizzo2022makedense}: treats the cylindrical grid as a
2-D image ($N_r$ range channels, $(\theta,z)$ spatial) and uses a
3-level circular-padded UNet (1.96M params).
\textbf{LMSCNet}~\cite{roldao2020lmscnet} (Cartesian grid): max-over-$z$
top-down BEV $\to$ stride-2 2-D encoder $\to$ 3-D decoder at $\tfrac{1}{4}$
scale $\to$ trilinear upsample (0.44M params, multi-scale auxiliary loss).

\begin{table}[H]
  \caption{FOSCo-Net (Gated Skip) vs.\ dense baselines (worst to best, $\tau{=}0.5$).
           $^\ddagger$Dense-input, no sparse encoder.
           Column defs and colour scale as Table~\ref{tab:ablation}.
           SFPR, the safety-critical traversability metric, is reduced
           by a factor of 2.1 compared with MIDUNet2D.}
  \label{tab:baselines}
  \centering
  \setlength{\tabcolsep}{3pt}
  \footnotesize
  \begin{tabular}{ll r rr r r}
    \toprule
    \textbf{Model} & \textbf{Grid}
      & \textbf{SFPR$\downarrow$}
      & \textbf{IoU$\uparrow$} & \textbf{F$_1\uparrow$}
      & \textbf{Rec$\uparrow$}
      & \textbf{Prec$\uparrow$} \\
    \midrule
    MIDUNet2D$^\ddagger$~\cite{vizzo2022makedense} & Cyl.  & \cc{0}0.741 & \cc{61}0.157 & \cc{62}0.268 & \cc{100}\textbf{0.710} & \cc{35}0.159 \\
    LMSCNet$^\ddagger$~\cite{roldao2020lmscnet}   & Cart. & $-$ & \cc{0}0.119  & \cc{0}0.210  & \cc{34}0.692 & \cc{0}0.141 \\
    \midrule
    \multirow{2}{*}{Ours (FOSCo-Net)}
      & Cyl.  & \cc{100}\textbf{0.346} & \cc{86}0.172 & \cc{86}0.290 & \cc{50}0.696 & \cc{89}0.187 \\
      & Cart. & $-$ & \cc{100}\textbf{0.181} & \cc{100}\textbf{0.303} & \cc{0}0.683 & \cc{100}\textbf{0.193} \\
    \bottomrule
  \end{tabular}
\end{table}

The results on the completion of the environment model are presented in Tab.~\ref{tab:baselines}. FOSCo-Net (Gated Skip) consistently outperforms both baselines.
Compared to MIDUNet2D~\cite{vizzo2022makedense}, FOSCo-Net~\cite{roldao2020lmscnet} improves precision by $+17.6\%$
($0.187$ vs.\ $0.159$), all-voxel IoU by $+9.6\%$ ($0.172$ vs.\ $0.157$),
and F$_1$ by $+8.2\%$ ($0.290$ vs.\ $0.268$), while sacrificing only $-2\%$
in stem recall ($0.696$ vs.\ $0.710$).
The SFPR gap is substantially larger: FOSCo-Net blocks only $35\%$ of the
traversable corridor versus $74\%$ for MIDUNet2D, a $2.1{\times}$
improvement that directly translates to safer path planning.
While MIDUNet2D leads on recall ($\mathrm{Rec}{=}0.710$) yet SFPR $0.741$
blocks 74\% of the traversable corridor, rendering it unsafe for path planning.
LMSCNet achieves competitive recall ($0.692$) but the lowest precision
($0.141$): its BEV-then-3D decoder recovers trunks by broad over-prediction.

\begin{table}[H]
  \caption{FOSCo-Net ablation on 10 held-out testing samples ($\tau{=}0.5$).
           All eight variants (four decoders $\times$ two grids); ordered
           worst to best by all-voxel accuracy.
           Colour scale: \cc{100}best \cc{0}worst per column.
           Rec~=~stem recall; Prec~=~global precision;
           SFPR~=~cylindrical-only shadow FPR (lower is better).
           \textbf{Bold}: best per column
           (SFPR: recall-competitive models, $\mathrm{Rec}_{\text{stm}}{>}0.65$).
           $^\dagger$Implicit: batch size 1.}
  \label{tab:ablation}
  \centering
  \setlength{\tabcolsep}{3pt}
  \footnotesize
    \begin{tabular}{ll r rr r r}
    \toprule
    & & \multicolumn{1}{c}{\textit{Trav.}}
      & \multicolumn{2}{c}{\textit{All-voxel}}
      & \multicolumn{1}{c}{\textit{Stem}}
      & \multicolumn{1}{c}{\textit{Pred. qual.}} \\
    \cmidrule(lr){3-3}\cmidrule(lr){4-5}\cmidrule(lr){6-6}\cmidrule(lr){7-7}
    \textbf{Architecture} & \textbf{Grid}
      & \textbf{SFPR$\downarrow$}
      & \textbf{IoU$\uparrow$} & \textbf{F$_1\uparrow$}
      & \textbf{Rec$\uparrow$}
      & \textbf{Prec$\uparrow$} \\
    \midrule
    \multirow{2}{*}{Implicit$^\dagger$}
      & Cyl.  & \cc{66}0.344 & \cc{39}0.161 & \cc{40}0.276 & \cc{37}0.508 & \cc{0}0.164 \\
      & Cart. & $-$ & \cc{0}0.149  & \cc{0}0.257  & \cc{19}0.453 & \cc{30}0.182 \\
    \midrule
    \multirow{2}{*}{Full Sparse}
      & Cyl.  & \cc{100}0.336 & \cc{50}0.165 & \cc{54}0.282 & \cc{3}0.405  & \cc{100}\textbf{0.226} \\
      & Cart. & $-$ & \cc{63}0.169 & \cc{66}0.288 & \cc{0}0.397  & \cc{94}0.222 \\
    \midrule
    \multirow{2}{*}{Dense Dec.}
      & Cyl.  & \cc{0}0.359 & \cc{64}0.170 & \cc{64}0.287 & \cc{91}0.668 & \cc{33}0.184 \\
      & Cart. & $-$ & \cc{70}0.171 & \cc{71}0.290 & \cc{89}0.664 & \cc{33}0.184 \\
    \midrule
    \multirow{2}{*}{Gated Skip}
      & Cyl.  & \cc{59}\textbf{0.346} & \cc{73}0.172 & \cc{71}0.290 & \cc{100}\textbf{0.696} & \cc{38}0.187 \\
      & Cart. & $-$ & \cc{100}\textbf{0.181} & \cc{100}\textbf{0.303} & \cc{95}0.683 & \cc{47}0.193 \\
    \bottomrule
  \end{tabular}
\end{table}

\subsection{Ablation Study}

The results of ablation study is presented in Tab.~\ref{tab:ablation} where we compare four presented architectures and two proposed grid models (cylindrical and cartesian).
Full Sparse under-predicts stems ($\mathrm{Rec}{\approx}0.40$): transposed
convolutions reach only cells within kernel distance of observed voxels,
missing far-field shadows.
Implicit over-predicts globally: without spatial convolutions it floods the
grid, giving the lowest precision ($0.164$--$0.182$) and Cartesian IoU
($0.149$) of all architectures. FOSCo-Net (Gated Skip) achieves the best all-voxel IoU/F$_1$ \\ (Cartesian: $0.181$/$0.303$), stem recall ($0.696$, Cyl.), and SFPR among recall-competitive architectures ($0.346$, Cyl.), outperforming the FOSCo-Net Dense Decoder by $+2$--$3\%$ in stem recall on both grids.
Grid choice has a smaller effect than decoder design: cylindrical and Cartesian variants differ by at most 2\% on most metrics, with Cartesian leading accuracy and cylindrical leading recall and traversability. A key structural difference between FOSCo-Net and both baselines is the
shared sparse encoder: by restricting convolutions to the
${\approx}0.7\%$ occupied voxels, it computes features exclusively where
LiDAR returns exist, producing encoder activations that are spatially
anchored to real surface observations. FOSCo-Net (1.90M params) achieves better accuracy with a comparable parameter budget.

\subsection{Qualitative Results}

\begin{figure}[htb]
  \centering
  \includegraphics[width=\linewidth]{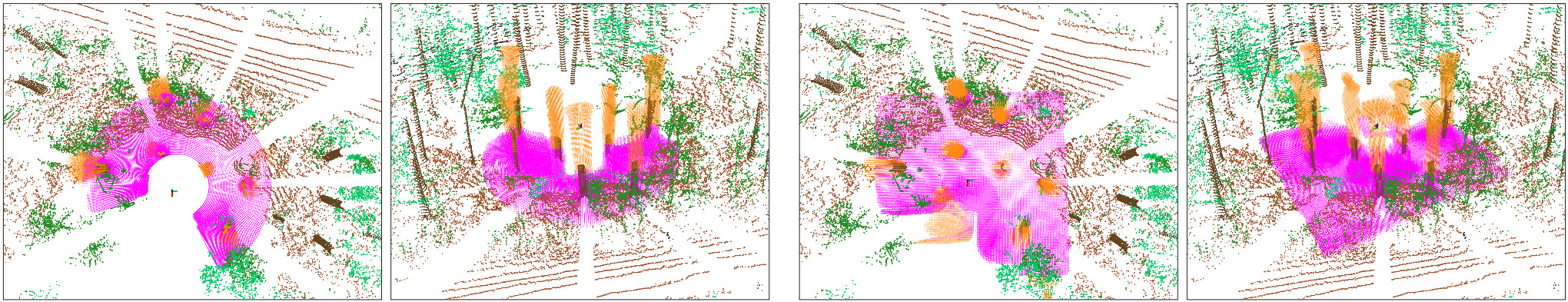}
  \put(-345,60){a} \put(-259,60){b} \put(-169,60){c} \put(-83,60){d}
  \caption{FOSCo-Net (Gated Skip) predictions overlaid on the input scan (a,c)
           (magenta\,=\,predicted ground, orange\,=\,predicted stem): cylindrical grid (a,b) and Cartesian grid (c,d).}
  \label{fig:qualitative}
\end{figure}
Fig.~\ref{fig:qualitative} shows FOSCo-Net (Gated Skip) predictions on
both grids for the same scene.
Both recover elongated occlusion shadows aligned with visible trunks,
confirming FOSCo-Net has learned to associate a stem at a given spatial
sector with a shadow extending behind it.
\begin{figure}[htb]
  \centering
  \includegraphics[width=\linewidth]{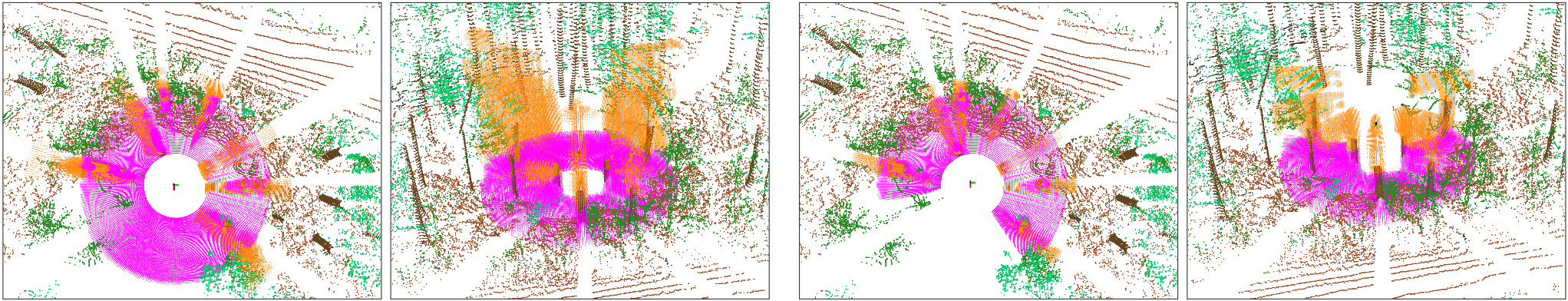}
  \put(-345,60){a} \put(-259,60){b} \put(-169,60){c} \put(-83,60){d}
  \caption{Dense baseline predictions, same scene as Fig.~\ref{fig:qualitative}
           (magenta\,=\,predicted ground, orange\,=\,predicted stem):Make-it-Dense (MIDUNet2D)~\cite{vizzo2022makedense} (a,b) and LMSCNet\cite{roldao2020lmscnet} (c,d). For both methods trunk locations are recovered but predictions smear across neighbouring voxels, consistent with the low precision ($0.141$) in
           Tab.~\ref{tab:baselines}.}
  \label{fig:qualitative_baselines}
\end{figure}
Fig.~\ref{fig:qualitative_baselines} shows the two failure modes from Table~\ref{tab:baselines}: MIDUNet2D floods wide angular sectors (SFPR\,=\,0.741) while LMSCNet smears voxels (precision $0.141$), both contrasting sharply with the tight occlusion-shadow geometry of FOSCo-Net (Gated Skip).

\section{Conclusion}
\label{sec:conclusion}

In this paper, we present FOSCo-Net, a self-supervised framework for predicting occluded forest geometry, ablated across two ego-centric grids and four decoder architectures. FOSCo-Net Cylindrical Gated-Skip ($\mathrm{SFPR}{=}0.346$) is the recommended model. In the future, we plan to extend the dataset to multiple sequences and
forest types for a larger-scale evaluation, to apply the proposed map
enhancement method to motion planning for an agricultural robot, and to
extend the Gated Skip decoder with angular attention and
uncertainty-aware predictions.

\subsubsection{Acknowledgements.}
Research conducted under the Applied Doctorate Programme of the Polish
Ministry of Science and Higher Education, 2024--2028
(grant No.\ DWD/8/0080/2024).

\bibliographystyle{splncs04}
\bibliography{references}

@misc{roldao2020lmscnet,
      title={LMSCNet: Lightweight Multiscale 3D Semantic Completion}, 
      author={Luis Roldão and Raoul de Charette and Anne Verroust-Blondet},
      year={2020},
      eprint={2008.10559},
      archivePrefix={arXiv},
      primaryClass={cs.CV},
      url={https://arxiv.org/abs/2008.10559}, 
}

@inproceedings{malladi2025digiforests,
  author    = {Meher V.R. Malladi and Nived Chebrolu and Irene Scacchetti and Luca Lobefaro and Tiziano Guadagnino and Beno\^{i}t Casseau
    and Haedam Oh and Leonard Frei{\ss}muth and Markus Karppinen and Janine Schweier and Stefan Leutenegger and Jens Behley
    and Cyrill Stachniss and Maurice Fallon},
  title     = {{DigiForests: A Longitudinal LiDAR Dataset for Forestry Robotics}},
  journal   = {Proc.~of the IEEE Intl. Conf. on Robotics \& Automation (ICRA)},
  year      = {2025},
}

@inproceedings{lin2017focal,
  author    = {Lin, Tsung-Yi and Goyal, Priya and Girshick, Ross and He, Kaiming and Doll{\'a}r, Piotr},
  title     = {Focal Loss for Dense Object Detection},
  booktitle = {Proceedings of the IEEE International Conference on Computer Vision (ICCV)},
  pages     = {2980--2988},
  year      = {2017},
  doi       = {10.1109/ICCV.2017.324}
}

@article{vizzo2022makedense,
  author  = {I. Vizzo and B. Mersch and R. Marcuzzi and L. Wiesmann and J. Behley and C. Stachniss},
  title   = {{Make it Dense: Self-Supervised Geometric Scan Completion of Sparse 3D LiDAR Scans in Large Outdoor Environments}},
  journal = {IEEE Robotics and Automation Letters (RA-L)},
  year    = 2022
}

@ARTICLE{yan2020sparsity,
  author={Yan, Lin and Liu, Kai and Belyaev, Evgeny},
  journal={IEEE Access}, 
  title={Revisiting Sparsity Invariant Convolution: A Network for Image Guided Depth Completion}, 
  year={2020},
  volume={8},
  number={},
  pages={126323-126332},
  doi={10.1109/ACCESS.2020.3008404}}

@ARTICLE{zioulis2022hybrid,
  author={Zioulis, Nikolaos and Albanis, Georgios and Drakoulis, Petros and Alvarez, Federico and Zarpalas, Dimitrios and Daras, Petros},
  journal={IEEE Access}, 
  title={Hybrid Skip: A Biologically Inspired Skip Connection for the UNet Architecture}, 
  year={2022},
  volume={10},
  number={},
  pages={53928-53939},
  doi={10.1109/ACCESS.2022.3175864}}

@inproceedings{xujia2025,
  title={Octree Diffusion for Semantic Scene Generation and Completion},
  author={Xujia Zhang and Brendan Crowe and Christoffer Heckman},
  year={2025},
  url={https://api.semanticscholar.org/CorpusID:281420372}
}

@Article{liang2024,
AUTHOR = {Liang, Li and Akhtar, Naveed and Vice, Jordan and Mian, Ajmal},
TITLE = {Voxel- and Bird’s-Eye-View-Based Semantic Scene Completion for LiDAR Point Clouds},
JOURNAL = {Remote Sensing},
VOLUME = {16},
YEAR = {2024},
NUMBER = {13},
ARTICLE-NUMBER = {2266},
URL = {https://www.mdpi.com/2072-4292/16/13/2266},
ISSN = {2072-4292},
DOI = {10.3390/rs16132266}
}

@INPROCEEDINGS{mei2023,
  author={Mei, Jianbiao and Yang, Yu and Wang, Mengmeng and Huang, Tianxin and Yang, Xuemeng and Liu, Yong},
  booktitle={2023 IEEE/RSJ International Conference on Intelligent Robots and Systems (IROS)}, 
  title={SSC-RS: Elevate LiDAR Semantic Scene Completion with Representation Separation and BEV Fusion}, 
  year={2023},
  volume={},
  number={},
  pages={1-8},
  doi={10.1109/IROS55552.2023.10341742}}

@ARTICLE{gu2024,
  author={Gu, Shuo and Lu, Jiacheng and Yang, Jian and Xu, Cheng-Zhong and Kong, Hui},
  journal={IEEE Transactions on Intelligent Vehicles}, 
  title={Dense Top-View Semantic Completion With Sparse Guidance and Online Distillation}, 
  year={2024},
  volume={9},
  number={1},
  pages={481-491},
  doi={10.1109/TIV.2023.3268241}}

@INPROCEEDINGS{Cao2025,
  author={Cao, Helin and Behnke, Sven},
  booktitle={2025 IEEE/RSJ International Conference on Intelligent Robots and Systems (IROS)}, 
  title={{DiffSSC}: Semantic {LiDAR} Scan Completion using Denoising Diffusion Probabilistic Models}, 
  year={2025},
  pages={2185-2192},
  doi={10.1109/IROS60139.2025.11246951}}

@InProceedings{Cheng2021,
  title = 	 {{S3CNet}: A Sparse Semantic Scene Completion Network for LiDAR Point Clouds},
  author =       {Cheng, Ran and Agia, Christopher and Ren, Yuan and Li, Xinhai and Bingbing, Liu},
  booktitle = 	 {Proceedings of the 2020 Conference on Robot Learning},
  pages = 	 {2148--2161},
  year = 	 {2021},
  editor = 	 {Kober, Jens and Ramos, Fabio and Tomlin, Claire},
  volume = 	 {155},
  series = 	 {Proceedings of Machine Learning Research},
  month = 	 {16--18 Nov},
  publisher =    {PMLR},
  url = 	 {https://proceedings.mlr.press/v155/cheng21a.html}
}

@INPROCEEDINGS{Tao2022,
  author={Tao, Yifu and Popović, Marija and Wang, Yiduo and Digumarti, Sundara Tejaswi and Chebrolu, Nived and Fallon, Maurice},
  booktitle={2022 IEEE/RSJ International Conference on Intelligent Robots and Systems (IROS)}, 
  title={{3D} Lidar Reconstruction with Probabilistic Depth Completion for Robotic Navigation}, 
  year={2022},
  pages={5339-5346},
  doi={10.1109/IROS47612.2022.9981531}
  }

@INPROCEEDINGS{Boettcher2023,
  author={Boettcher, Wolfgang and Hoyer, Lukas and Unal, Ozan and Li, Ke and Dai, Dengxin},
  booktitle={2023 IEEE/RSJ International Conference on Intelligent Robots and Systems (IROS)}, 
  title={{LiDAR} Meta Depth Completion}, 
  year={2023},
  pages={7750-7756},
  doi={10.1109/IROS55552.2023.10341349}
  }

@misc{zhong2023shine,
      title={SHINE-Mapping: Large-Scale 3D Mapping Using Sparse Hierarchical Implicit Neural Representations}, 
      author={Xingguang Zhong and Yue Pan and Jens Behley and Cyrill Stachniss},
      year={2023},
      eprint={2210.02299},
      archivePrefix={arXiv},
      primaryClass={cs.CV},
      url={https://arxiv.org/abs/2210.02299}, 
}

@inproceedings{zhong2024ndf,
  author = {Xingguang Zhong and Yue Pan and Cyrill Stachniss and Jens Behley},
  title = {{3D LiDAR Mapping in Dynamic Environments using a 4D Implicit Neural Representation}},
  booktitle = {{Proc. of the IEEE/CVF Conf. on Computer Vision and Pattern Recognition (CVPR)}},
  year = 2024,
}

\end{document}